\documentclass[journal]{IEEEtran}

\usepackage{booktabs}
\usepackage{url}
\usepackage{cite}

\ifCLASSINFOpdf
\usepackage{graphicx}
\else
\fi

\begin{document}

\title{Enhancing Human Action Recognition and Violence Detection Through Deep Learning Audiovisual Fusion\thanks{This work has been submitted to the IEEE for possible publication. Copyright may be transferred without notice, after which this version may no longer be accessible.}}
\author{Pooya~Janani,
        Amirabolfazl~Suratgar,
        and~Afshin~Taghvaeipour,
\thanks{P.Janani: Research Assistant, Distributed and Intelligent Optimization Research Laboratory, Dept. of Electrical Engineering, Amirkabir University of Technology, Tehran, Iran.}
\thanks{A. A. Suratgar: Associate Professor, Distributed and Intelligent Optimization Research Laboratory, Dept. of Electrical Engineering, Amirkabir University of Technology, Tehran, Iran. e-mail (Corresponding author): \\
(a-suratgar@aut.ac.ir).}
\thanks{A. Taghvaeipour: Associate Professor, Dept. of Mechanical Engineering, Amirkabir University of Technology, Tehran, Iran.}
}

\markboth{}%
{  }

\maketitle

\begin{abstract}
This paper proposes a hybrid fusion-based deep learning approach based on two different modalities, audio and video, to improve human activity recognition and violence detection in public places. To take advantage of audiovisual fusion, late fusion, intermediate fusion, and hybrid fusion-based deep learning (HFBDL) are used and compared. Since the objective is to detect and recognize human violence in public places, Real-life violence situation (RLVS) dataset is expanded and used. Simulating results of HFBDL show 96.67\% accuracy on validation data, which is more accurate than the other state-of-the-art methods on this dataset. To showcase our model's ability in real-world scenarios, another dataset of 54 sounded videos of both violent and non-violent situations was recorded. The model could successfully detect 52 out of 54 videos correctly. The proposed method shows a promising performance on real scenarios. Thus, it can be used for human action recognition and violence detection in public places for security purposes. 
\end{abstract}


\IEEEpeerreviewmaketitle

\section{Introduction}

\IEEEPARstart{H}{uman} action recognition refers to the process of identifying an action through the use of a system, usually artificial and computational intelligence systems, to learn about the action and then applies that knowledge to detect similar actions \cite{Ramanathan2014}. One important category of these actions is violence. The team of author’s research laboratory has done some research in in computational intelligence application and their applications \cite{MALAKOUTI2024100278, MALAKOUTI2023102337, MIRZABAGHERIAN2023107159, 9663528, 5478900}. In recent years, the rise in violent incidents in public places has heightened concerns over public safety. As a result, surveillance cameras have been extensively deployed at various locations \cite{Bhatt2021}. However, the ever-increasing number of cameras has led to a proportional increase in operators and supervisors required to monitor video feeds \cite{Lejmi2018}, which poses significant challenges and limitations on real-time monitoring. Since crowd monitoring has increasingly been as one of the research interests and application areas, it is important to achieve a human-level performance; therefore, A large number of training data sets and effective algorithms are required \cite{Perera2020}. Due to this fact, developing automated video surveillance systems (VSSs) has become crucial to ensure the security and safety of the population \cite{Himeur2023}. One application of these systems is to recognize action and detect violence. Researchers have developed various techniques for violence detection and human action recognition using either unsupervised \cite{Hadikhani2023, Qi2020} or supervised learning. Among supervised techniques, one approach is using video-only models \cite{Senst2017,Soliman2019, Irfanullah2022, GARCIACOBO2023SkeletonsViolence, s24020668, Matei2022, Haroon2022, Ullah2023}. The other approach is to use audio-only models \cite{BAKHSHI2023244, Reinolds2022, Santos2021, duraes2023violence, Yıldız2023}. One of the promising approaches which has recently been of significant importance is to fuse data from multiple modalities \cite{JAAFAR2023118523, anwar2022deepsafety}, such as video and audio, to enhance the accuracy of detection systems. although visual violence recognition has more specific and practical value and it is therefore very significant to make investigations \cite{Sun2019}, research has demonstrated that combining auditory and visual data can lead to more sophisticated outcomes in various fields. For instance, one study \cite{Oh2019} demonstrated a novel approach for facial reconstruction using recorded voice, while another \cite{Ginosar2019} explored the conversion of speech signals into gestures. These findings highlight the strong correlation between auditory and visual information, which can complement each other in many ways. In particular, audio information has been found to play a crucial role in violence detection projects. For example, during fights or gun battles, the sound of shouting, profanity, or gunfire can be detected from significant distances, making it a valuable tool for identifying violent incidents. Therefore, the integration of audio and visual information has great potential for improving our understanding of complex events and developing more effective solutions for addressing them.

In recent years, a number of studies have been conducted on the field of human activity recognition and violence detection. We present a few of them in this article that we believe are pertinent to the framework of this study.
Violence detection was carried out by researchers in \cite{Gracia2015} step-by-step. They started by removing the background from the video frames, leaving only the moving parts of each frame—referred to as motion bubbles—in each frame. The authors assumed that these motion blobs have a particular shape and position. The K largest are ultimately categorized as either violent or nonviolent after analysis. This method uses an ellipse detection method to detect objects. Also, an algorithm to find the acceleration was used to extract features, and finally spatiotemporal features were used in an effort to classify the scenes. This method was tested with both crowded and less-crowded scenes, yielding a near 90\% rate of accuracy.
In a study \cite{Peixoto2018}, violence-related features like blood, explosions, and fights were learned using separate networks. Afterwards, distinct SVM classifiers were trained for each of these concepts to describe violence using these features. The results from the classifiers were then combined into a meta-classification. Using same dataset, in \cite{Peixoto2020} researchers used independent networks for multimodal violence detection using video and audio. They used a late fusion approach, where the decision is made based on both the visual and audio features. The proposed method has potential applications in security and surveillance systems, where it can be used to detect violent behavior in real-time.
In \cite{Reinolds2022}, violence detection was performed using separate audio and video models on the RLVS dataset. Since some videos in the dataset lacked sound, the authors up-dated the dataset to include 250 videos for each class. The results showed that the visual model achieved higher accuracy (89\%) than the audio model (76\%), indicating that the visual model is more effective at identifying violent behavior in videos on this dataset.
It is worth mentioning that data fusion can be done by different modalities. For example, in \cite{chen2015} two modalities from a Kinect depth camera and a wearable inertial body sensor (accelerometer) was used with the help of data fusion for the purpose of human action recognition, or in \cite{Li2019} to benefit from both LSTM and CNN models, multiply fusion was used to utilize different views of data generated from a skeleton sequence. In \cite{Taghanaki2023} a fusion module is used as well to integrate the features extracted from accelerometer signal in time domain and localized time-frequency domain.

In this paper, we aim to investigate different fusion strategies and evaluate their performance in activity recognition and violence detection. Our proposed method analyzes both visual and audio cues to detect potential threats. The final model with the best accuracy will be used in an interactive robot that will be placed in public places like airports and one of its goals is to increase public security through continuous monitoring of the surrounding environment through the corresponding video and sound and giving an early intimation to the authorized person when needed. This project has the potential to contribute to developing more efficient and reliable surveillance systems for public safety. The present study makes a significant contribution to the field of violence detection through audio-visual fusion by expanding and modifying the Real-Life Violence Situations (RLVS) dataset \cite{Soliman2019}. We aimed to create a diverse dataset that includes both violent and non-violent conditions, captured in various public situations and places. Notably, we ensured that all videos in the final dataset include relevant sound, which is an essential feature for the project.
In terms of the article's structure, the article begins with an introduction containing the big picture and literature review, followed by an explanation of the methodology used in the study, including the materials and methods. The results are then presented and analyzed before drawing conclusions.

\section{MATERIALS AND METHODS}
To conduct this research, Python and some of its relevant libraries and also TensorFlow framework are used to create and train the models. It is possible to build an entire pipeline for the development, training, and evaluation of models using these tools, as detailed in the next subsection. Each of the fusion strategies used for fusing audio and video features, which are extracted in parallel from pretrained models, are tested using different settings in order to understand which configuration better suited each strategy and pretrained model.
Data is provided to pretrained video and audio models during the training phase, with the aim of extracting features from them and then fusing this knowledge based on different scenarios in order to understand whether it is violent or not. Data augmentation is used for this stage in order to enhance the model generality and increase its accuracy on validation stage. 

\subsection{Model Implementation}
This subsection aims to provide a better comprehension of the fusion model discussed by presenting the research pipeline in Fig.  1 This graphical representation illustrates how information is processed and the steps taken for training and evaluating the models. 
The first step involves loading the dataset and selecting a batch of videos based on the defined hyperparameter called batch size (BS). The audio and video data of these selected videos are extracted, and Mel spectrograms are generated from their audio data. Similarly, the video frames are extracted and loaded into a Tensor to enable feature extraction by pre-trained models.
In the subsequent steps, both pre-trained models receive their respective data and extracted features to create a higher-level representation. The resulting representations of the video and audio data are then fed into the fusion module. Finally a classification decision is made based on the combination of the two modalities.

\begin{figure}[ht]
\centering
\includegraphics[width=0.48\textwidth]{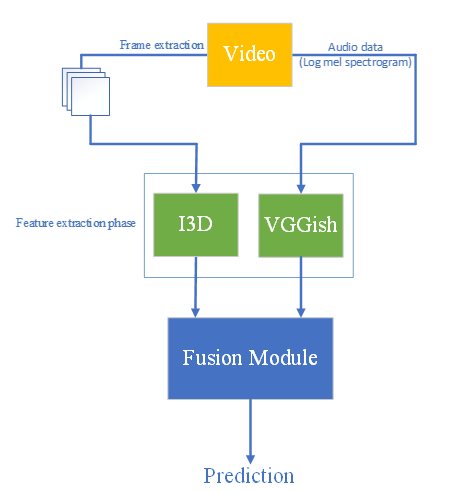}
\caption{System pipeline, which consists of 3 main phases: feature extraction, data fusion and prediction.}
\label{fig:overall4}
\end{figure}

\subsection{Data}
\begin{figure}[ht]
\centering
\includegraphics[width=0.48\textwidth]{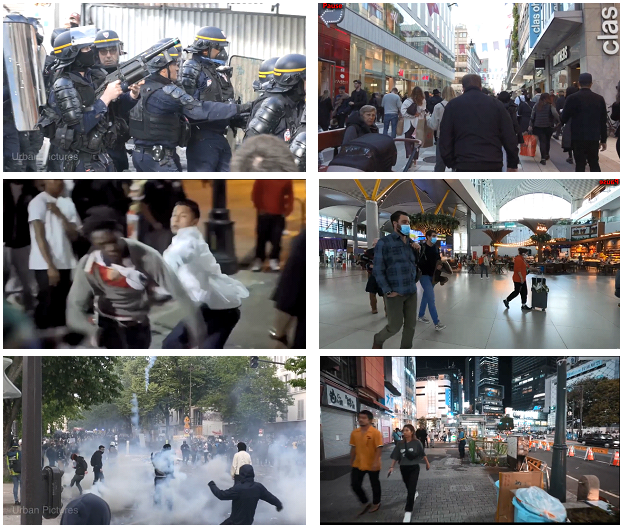}
\caption{Screenshots of the dataset's videos. (left column): Violent videos and (right column) non-violent videos.}
\label{fig:fig2}
\end{figure}

Since the final model would be used for real time security purposes in a public place, RLVS dataset, a dataset of real-life situations of violence and Nonviolence actions, is selected. As Mentioned in \cite{Reinolds2022}, It was discovered that the majority of videos in the dataset either lacked sound or contained sound that was unrelated or not applicable. After reviewing the dataset, it was discovered that 175 violent and 150 non-violent videos had relevant sound. However, due to an insufficient number of videos available for training the model, an attempt is made to look across the internet for relevant sound copies of the same videos, which are then substituted for the silent versions. To further expand the dataset, various videos from public areas are also collected with the ultimate goal of using the model in an interactive robot to monitor airport environments. The non-violent videos included footage of streets and alleys in different cities  
worldwide, both secluded and crowded with pedestrians. These videos were sourced from "city walking" videos filmed by individuals who simply walked around and captured footage of streets and alleys without spoken commentary. Moreover, videos of airport environments from around the world, such as Turkey and Dubai airports, are also collected. The videos are sourced from "airport tour" videos that do not include any spoken commentary and filmed by individuals walking around and capturing footage of various areas within airports around the world. 
In the category of violent videos, in addition to videos of conflict between individuals, to provide complete coverage of violent events that may occur in an airport or public place, such as fights, armed conflicts, riots, and creating terror, videos of violent protests by the people of France and other countries are added to the dataset. These videos contain scenes of fires, shootings, riots and so on. It should be noted that all of the added videos have relevant sound and every effort has been made to include various spaces and locations to provide diversity in the dataset.
Finally, to have a balanced dataset, 300 violent videos, and 300 non-violent videos are collected. All of these videos depict real-life scenarios and events in public places such as streets, alleys, and other covered public areas including airports. Fig. \ref{fig:fig2} displays several examples of the dataset scenes.

\begin{figure}[ht]
\centering
\includegraphics[width=0.48\textwidth]{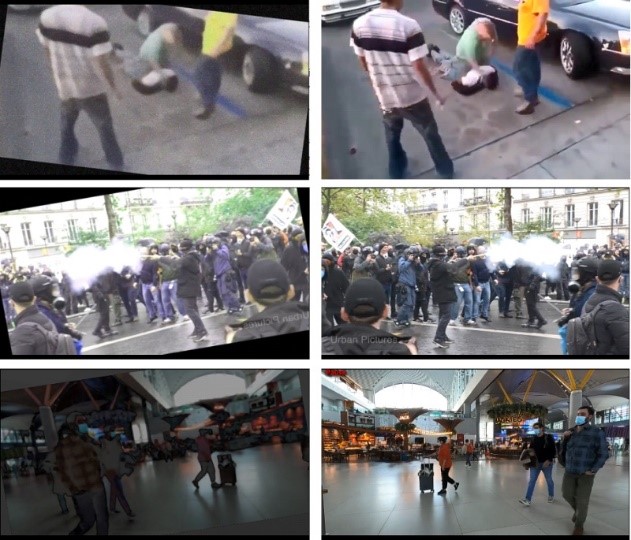}
\caption{Screenshots of videos in the augmented dataset. (left column): Augmented videos and (right column): original videos.}
\label{fig:fig3}
\end{figure}

To enhance the accuracy of model on unseen data and improve its generalization ability, data augmentation technique is employed. This involved creating modified versions of the input samples to artificially expand the size of the training dataset. The augmentation is particularly useful when dealing with small datasets as it enables more diverse representations of the target concept to be provided. For this purpose, the sound and video components of each video data is separated, and various data augmentation methods are randomly (in terms of the number and order of their application) applied to each component. By doing so, the amount of data in the dataset is increased (got tripled), and the ability of model to recognize violence across a wider range of inputs is improved. Some of results can be seen in Fig. \ref{fig:fig3}. 
The main focus of the data augmentation is to develop a multimodal model that is robust against variations in color, brightness, rotation, noise, and blurring of the video data and variations in sound pitch, volume, and some types of noise of the audio data. These functions and the modality the affect on, can be seen in Table \ref{tab:aug}.

\begin{table}[ht]
\centering
\caption{List of data augmentation techniques used}
\label{tab:aug}
\begin{tabular}{l l}
\toprule
Modality affected & Function name \\ \midrule
Video & Color Jittering\\
 & Rotation\\
 & Additive Noise\\
 & Horizontal/Vertical Flip\\
 & Gaussian Blur\\
  & Median Blur\\
   & Brightness/Contrast Adjustment \\
 \hline
Audio & Pitch Shift\\
 & Additive Noise\\
 & Volume Adjustment\\
  & Frequency Filtering \\
\bottomrule
\end{tabular}
\end{table}

\subsection{Data Preparation}
Two pretrained models, VGGish \cite{Hershey2017} and I3D \cite{Carreira2017} which will be discussed later in this paper, are used for feature extraction. In this regard, audio and video data should be extracted from videos, and data preprocessing should be conducted on audio and video data to be suitable for pretrained models.
To prepare the video data, several preprocessing steps need to be performed. The input video frames are first cropped to a square shape centered on the original frame and then resized to 224x224 to remove any unnecessary areas of the frames. Finally, the pixel values of each frame are normalized to fall within the range of 0 and 1, which aids in the training process of machine learning models. 
For audio data similar to \cite{Hershey2017}, first, all audio data is resampled to 16 kHz mono. Then, a spectrogram is created using Short-Time Fourier Transform with a window size of 25 ms, a window hop of 10 ms, and a periodic Hann window. The   spectrogram is transformed into a Mel spectrogram by mapping it to 64 Mel bins covering the range between 125 to 7500 Hz. Finally, these features are framed into non-overlapping examples of 0.96 seconds, including 64 Mel bands and 96 frames of 10 ms each. 

\subsection{Modelling}
For each modality, a pretrained model is needed to extract features. Then two modalities, high-level representation of audio and video, are fused through different strategies, namely, intermediate, late and hybrid fusion. In \cite{Reinolds2022} two pre-trained models, Resnet MC18 and Resnet (2+1) D \cite{Tran2018} were used for video classification part and the average validation accuracy of 89\% on 500 videos (250 for each class) of modified RLVS dataset were achieved. For audio classification the best result was obtained with Resnet18 \cite{He2016} which achieved the average validation accuracy of 76\% on 500 audio data of videos of modified RLVS dataset. In order to conduct further research on existing architectures of video and audio classifications and expand observations, the I3D network and VGGish network are also incorporated to classify the video and the audio data, respectively.

In the following subsections, the best-performing audio and video models are selected based on their accuracy on the modified RLVS dataset. The features extracted by these models are then integrated using a hybrid fusion approach, which combines the advantages of intermediate and late fusion.

\subsubsection{Audio model for feature extraction phase}
Machine learning techniques for analyzing audio involve extracting relevant information from a sound recording, such as the frequency and amplitude of the sound wave over time. One way to extract low-level representations of features is by computing the log-Mel spectrogram of the audio signal. This involves breaking the audio signal into small segments, called frames, and then applying the Short-Time Fourier Transform (STFT) to each frame to obtain its frequency content \cite{Purwins2019}. The resulting spectrogram is then processed to produce a log-magnitude representation, which emphasizes lower frequencies that are more important for human perception. As it was mentioned earlier, the audio model being used in this study is VGGish. This decision is made after using this model for audio classification of the modified RLVS dataset and the results, which will be discussed in the next section, are better than the model being used in \cite{Reinolds2022}. The VGGish architecture, which is shown in Table \ref{tab:vgg} is a convolutional neural network designed for audio-based applications such as speech recognition and audio classification. It works by converting audio signals into spectrograms, which are then fed as input to the VGGish model. The model consists of several layers that apply different operations on the input data. The first Conv2D layer applies 96 filters of size 3x3 to the input spectrogram and applies the Rectified Linear Unit (ReLU) activation function to the output. This generates 96 feature maps of size 64x64. Following the first Conv2D layer, a MaxPooling2D layer with a pool size of 2x2 reduces the spatial dimensions of the output while retaining essential features. Two more Conv2D layers follow, each with 128 and 256 filters of size 3x3, respectively. Again, a MaxPooling2D layer with a pool size of 2x2 follows each convolutional layer to reduce the feature map dimensions. Next, two Conv2D layers apply 512 filters of size 3x3 to the input spectrogram, generating 512 feature maps, each with size 12x8. As with the previous Conv2D layers, a MaxPooling2D layer reduces the feature map dimensions. 
\\ \indent Finally, the Flatten layer converts the output of the previous layer into a one-dimensional array that serves as input to the fully connected layers. The three dense layers contain different numbers of neurons which learn to classify the input spectrogram based on the extracted features. The ReLU activation function is applied to the outputs of the first two dense layers.

\begin{table}[ht]
\centering
\caption{VGGish architecture}
\label{tab:vgg}
\begin{tabular}{l l l l l}
\toprule
Layer & Output shape & Kernel size & Stride & Activation \\ \midrule
(Conv2D) & (96, 64, 64) & 3 x 3 & 1 & ReLU \\
(MaxPooling2D) & (48, 32, 64) & 2 x 2 & 2 & - \\
(Conv2D) & (48, 32, 128) & 3 x 3 & 1 & ReLU \\
(MaxPooling2D) & (24, 16, 128) & 2 x 2 & 2 & - \\
(Conv2D) & (24, 16, 256) & 3 x 3 & 1 & ReLU \\
(Conv2D) & (24, 16, 256) & 3 x 3 & 1 & ReLU \\
(MaxPooling2D) & (12, 8, 256) & 2 x 2 & 2 & - \\
(Conv2D) & (12, 8, 512) & 3 x 3 & 1 & ReLU \\
(Conv2D) & (12, 8, 512) & 3 x 3 & 1 & ReLU \\
(MaxPooling2D) & (6, 4, 512) & 2 x 2 & 2 & - \\
(Flatten) & 12288 & - & - & - \\
(Dense) & 4096 & - & - & ReLU \\
(Dense) & 4096 & - & - & - \\
(Dense) & 128 & - & - & - \\
\bottomrule
\end{tabular}
\end{table}

\begin{figure*}[ht]
\centering
\includegraphics[width=\textwidth]{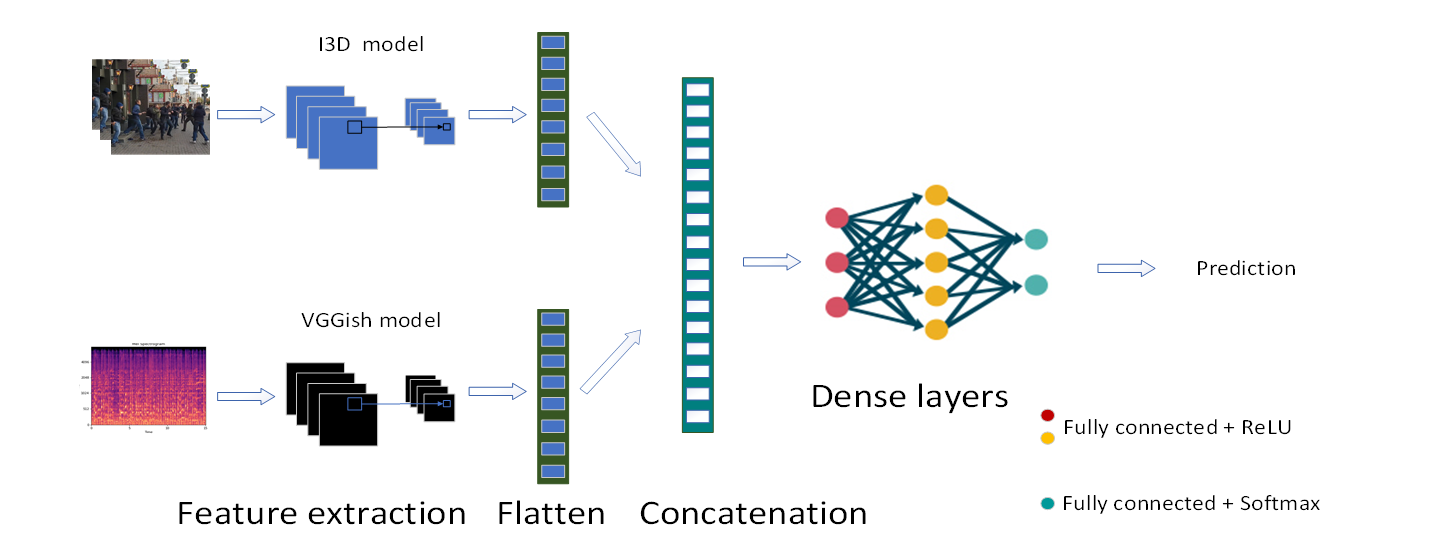}
\caption{Intermediate fusion model architecture. }
\label{fig:fig5}
\end{figure*}

\subsubsection{Video model for feature extraction phase}
Video analysis in Machine Learning (ML) has been a challenging task due to the high-dimensional nature of video data and the need for capturing temporal dependencies. In recent years, 3D Convolutional Neural Networks (3D CNNs) have emerged as a powerful tool for extracting features from videos. One popular variant of 3D CNNs is the Inflated 3D CNN, which known as I3D network, which is designed for video analysis tasks such as human action recognition. It is created by converting 2D convolutions to 3D convolutions using pre-trained Inception V1 networks and then trained on large-scale Kinetics Human Action Video Dataset \cite{Kay2017}, which is a dataset of human action videos that contains around 300,000 video clips covering 400 different action categories. The Inception V1 network is a widely used 2D CNN architecture that has been trained on the ImageNet \cite{Russakovsky2015} dataset. To convert this 2D CNN to a 3D CNN for video analysis, all the kernels in the pre-trained Inception V1 network are increased from two to three dimensions. This means that the weights and biases of the kernels are preserved, while an additional dimension is added to account for temporal information in videos. However, since the pre-trained 2D kernels cannot be directly applied to 3D data, the additional 3D dimension is initialized with random weights. This way, the network can learn to capture spatiotemporal features in videos while still leveraging the knowledge learned from the 2D CNN. By using this technique of converting 2D to 3D convolutions, the I3D network is able to transfer the features learned by the Inception V1 network on images to the domain of videos. This helps to improve the performance of the I3D network compared to training from scratch and makes it easier to be used as a pre-trained model for video analysis tasks such as violence detection, where there may not be enough labeled data to train a powerful model from scratch.

\subsubsection{Fusion module}
In order to classify videos in terms of the presence or absence of violence, the extracted audio and video should be integrated under a strategy. There are many application areas that require multimodal data integration, Such as biomedical applications, transportation systems, interactive robots, and etc. Audio and video analysis is a special case of multimodal analysis where the input sources are audio and video. These two types of data are usually related and convey complementary information. For example, seeing a person's face will help to understand his speech, or hearing the sound of what is happening will help to understand the type of event. In general, audio and video analysis consist of two main steps. In the first step, appropriate features are extracted from each of modalities, which were discussed previously, and then the extracted features are integrated. Integration of features can be conducted under three strategies \cite{Katsaggelos2015}, namely, early, intermediate and early fusion which will be explained in the sequel.

Early fusion involves combining the extracted features from different modalities (in this case, audio and video data) into a single feature representation before feeding it to the fusion model. By doing so, the model can leverage the correlated information across the different modalities at the feature level, which potentially improves the performance \cite{Katsaggelos2015}.
    
Intermediate fusion is similar to early integration, but with an added step. Each input data source undergoes feature extraction and is processed through multiple layers (e.g., 2-D-convolutions, 3-D-convolution, or fully connected) to obtain a higher-level representation of features before using a fusion layer \cite{Ramachandram2017}. This approach allows for more complex relationships between the inputs to be captured, particularly in applications like computer vision and sound event detection, where different modalities of information need to be combined. However, the effectiveness of this method depends on the specific application and quality of the input data. 

\begin{figure*}[ht]
\centering
\includegraphics[width=\textwidth]{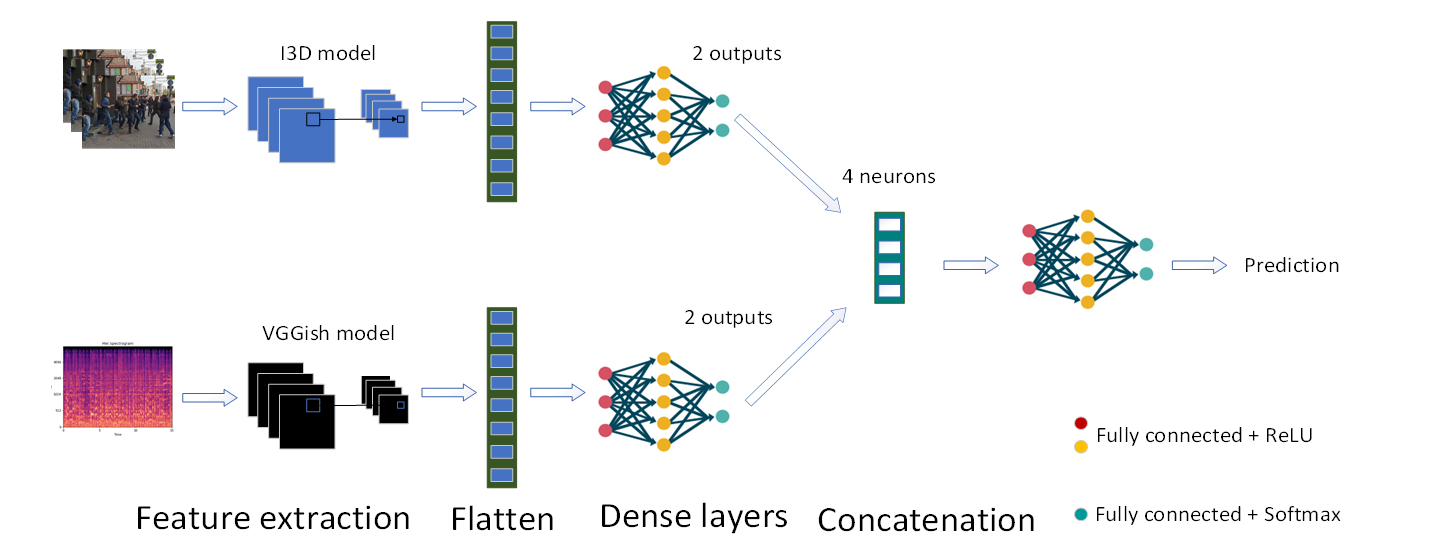}
\caption{Late fusion model architecture.}
\label{fig:fig6}
\end{figure*}
Late fusion involves processing audio and video data separately and then combining them at a later stage to make a final decision or prediction. In this approach, the separate modeling of audio and video data allows for the use of specific techniques for each type of data, which may improve the performance.

Since the extracted features are at a high level of representation of the modalities, to combine these features effectively, two strategies can be used: intermediate fusion and late fusion. These strategies involve different approaches to combine the features, and their effectiveness can be examined to determine which approach works better for a given task. To take advantage of both intermediate and late fusion, HFFBDL method is proposed. Accordingly, it effectively integrates the complementary strengths of both modalities, audio and video.

\subsubsection{Final model}

To implement the intermediate fusion, first the audio data and video data are separated from each other. Then, high-level representations of each modality are extracted separately using the I3D model, which is already trained on the Kinetics dataset for the video data, and the pre-trained VGGish model for the audio data. Finally, the extracted features are integrated at the beginning of the classification model and then passed through fully connected layers to make the prediction. Fig. \ref{fig:fig5} shows the architecture of the proposed model.

To implement the late fusion, the output of separate models that process audio and video data are first classified separately. The final decisions from both models are then combined or fused together and fed into a fully connected neural network, which determines the ultimate decision of the entire model. Fig. \ref{fig:fig6} depicts the proposed model architecture which uses this strategy of fusion.

\begin{figure*}[ht]
\centering
\includegraphics[width=\textwidth]{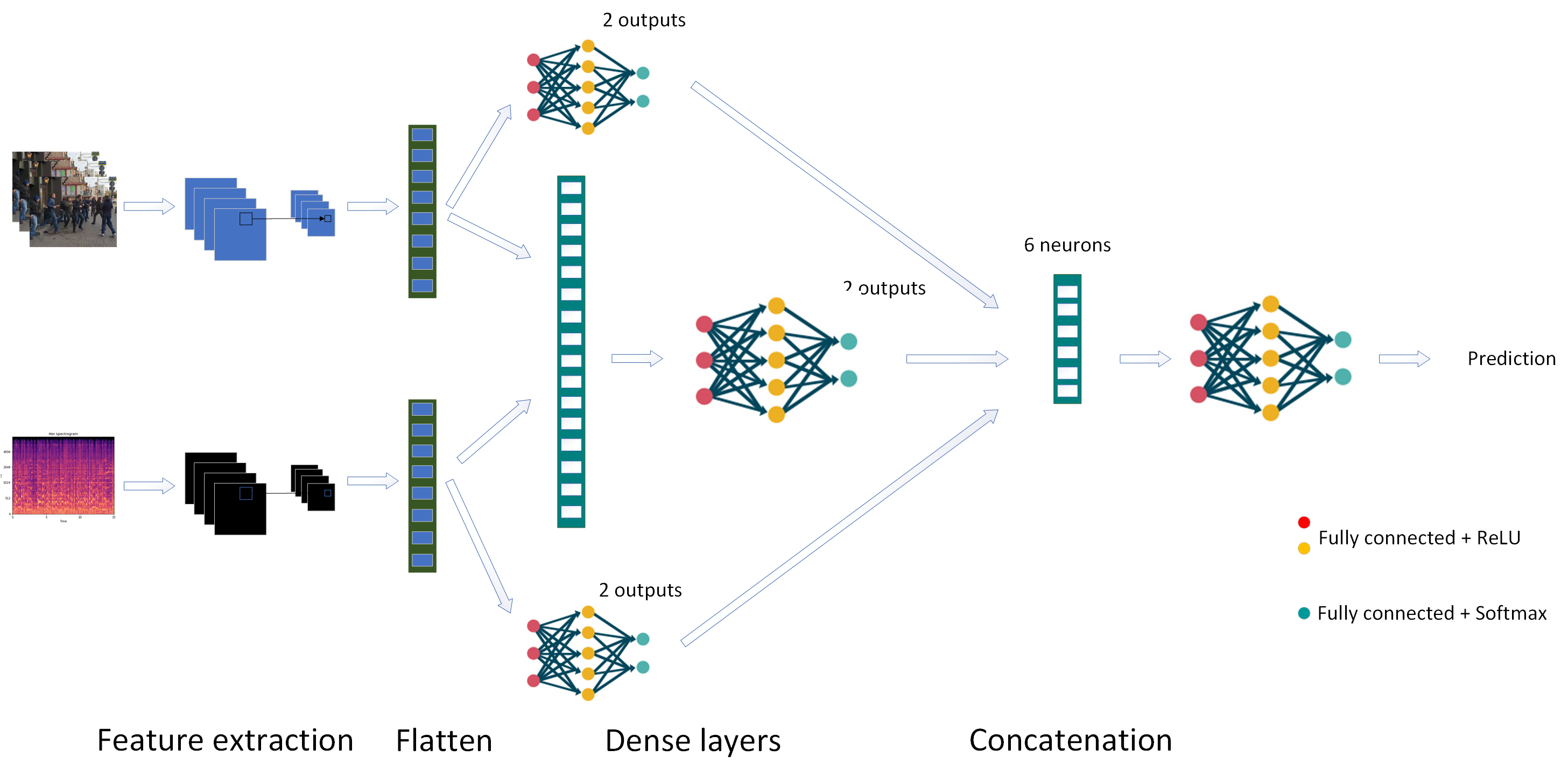}
\caption{Proposed hybrid fusion model architecture.}
\label{fig:fig7}
\end{figure*}

In the proposed implementation of HFBDL strategy, as illustrated in Fig. \ref{fig:fig7}, the combination of two integration methods, namely intermediate and late fusions, is employed. In this approach, the high-level features of both audio and video modalities are integrated at the onset of the classification model, reflecting the concept of intermediate fusion. Simultaneously, each modality is independently modeled. Eventually, the three outputs, originating from the audio model, video model, and intermediate fusion, are integrated to be classified by neural networks.

\subsection{Evaluation}

In order to evaluate each fusion strategy, several metrics are taken into account and examined. It is necessary to mention that the average is calculated through randomly splitting the dataset into train and validation sections:

\begin{itemize}
    \item \textbf{Average training accuracy(ATA):} The average accuracy percentage a fusion model can attain using the training data.
    \item \textbf{Average training loss(ATL):} The average loss a fusion model can attain using the training data.
    \item \textbf{Average validation accuracy(AVA):} The average accuracy percentage a fusion model can attain using the validation data.
    \item \textbf{Average validation loss(AVL):} The average loss a fusion model can attain using the validation data.
\end{itemize}

\section{RESULTS AND DISCUSSION}
For training each fusion approach, different hyperparameters and conditions are considered. Some of these hyperparameters are learning rate (LR), batch size (BS), number of dense layers, and number of neurons for each dense layer. For all of the approaches, the optimizer is Adam and since the task is classification, the loss function is set to categorical cross entropy. After every dense layer, a dropout layer is used, whose rate is another hyperparameter which is achieved via random search technique. After undergoing training under various conditions and adjusting different hyperparameters for each fusion strategy, a results set was obtained. Finally, based on the results, the dropout rate is set to 0.5, which means that during training, each neuron has a probability of being dropped out is 50\%. Also, LR and BS are set to 0.0001 and 7, respectively, for each method.

In the following subsections, first the two pretrained model will be examined separately and the results will show the reason of choosing them as the feature extraction models of audio and video data. Then, the results of the three fusion strategies will be shown and the best strategy will be selected as the final model for detecting violence in public places.

\subsection{Audio Pretrained Model}
The classification of audio to detect violent acts is based on the analysis of sound characteristics commonly associated with such events. These may include screaming, banging or battering of surfaces, breaking sounds, and other similar noises. The underlying concept behind this approach is that a classifier can be trained to recognize patterns in audio streams and identify when violent acts are taking place. This technology has the potential to provide valuable insights into different scenarios where violence may occur, including public disputes, public protests, and criminal activities. By leveraging machine learning algorithms and advanced signal processing techniques, audio classification systems can help law enforcement agencies and other organizations respond more effectively to incidents which involve violence. Overall, the development of these technologies represents an important step forward in our ability to detect, prevent, and address violent behavior in a variety of settings.

In Table \ref{tab:tab4}, audio classifier’s performance based on VGGish pretrained model can be seen. As it is clear, the results of the audio classifier based on the pretrained VGGish model are better than those of the audio classifiers in \cite{Peixoto2018}. Therefore, this model is recommended in the final fusion approach as it can learn audio patterns more effectively.

\subsection{Video Pretrained Model}
Video classifiers are currently at the forefront of research on violence detection, and they are expected to perform exceptionally well due to their ability to provide detailed visual information about violent actions. This is in contrast to audio classifiers, which have limitations in identifying violent actions based solely on the produced sounds. State-of-the-art papers in this field primarily focused on video classifiers rather than audio-based approaches.
One reason for the superior performance of video classifiers is that they can capture more descriptive features than audio classifiers. These models utilize visual cues such as motion, shape, and texture to identify specific actions and movements associated with violent behavior. Additionally, video classifiers can distinguish between different types of violent actions with a greater level of precision compared to audio classifiers, which may struggle to differentiate similar sounds. Therefore, utilizing video classifiers represents a promising approach for improving the accuracy of violence detection systems.

Looking at Table \ref{tab:tab4} it is evident that the I3d model outperforms Resnet(2+1)D model \cite{Reinolds2022} on the video dataset. This improvement in performance can be attributed to the fact that the I3d model is specifically designed and trained for action recognition tasks in videos, whereas the other model is not originally intended for this purpose.

\subsection{Effect of data augmentation on models}
To verify the effect of our augmentation method on the models' performance, both audio-only and video-only pre-trained models was firstly fine-tuned on the original dataset separately, and then the data augmentation was investigated to achieve better results on the AVA. This measure was taken to show the better performance of both VGGish and I3D models even on the non-augmented dataset. As shown in Table \ref{tab:aug_result}, it can be inferred that the whole augmentation method is effective; accordingly, this augmentation method is used in the final system pipeline to train models comprising fusion module.

\begin{table}[ht]
\centering
\caption{The percentage of the video-only and audio-only AVA on the augmented and non-augmented datasets.}
\label{tab:aug_result}
\begin{tabular}{l l l}
\toprule
Model Name & Before augmentation (\%) & After augmentation (\%)\\ \midrule
I3D & 89.50 & 91.00 \\
 \hline
VGGish & 78.00 & 80.00\\

\bottomrule
\end{tabular}
\end{table}

\subsection{Multimodal model based on three strategies of fusion} The fusion of multiple modalities, such as video and audio, has a significant impact on violence detection compared to detecting violence based solely on video. While a video file provides a detailed visual representation of violent actions, detecting violence using only video can be challenging in certain situations, for example, when actions are obscured or occur in low-light environments. Audio, on the other hand, can provide vital information about violent actions that may not be visible in the video, such as the sound of a punch or a gunshot. By combining video and audio inputs through multimodal fusion techniques, the resulting model can leverage both modalities' strengths, and it ended up with a more comprehensive understanding of violent actions. This enables the model to detect violence accurately in a wider range of scenarios, regardless of the visibility of actions in the video.

In Table \ref{tab:tab4} the first three rows show the results of the fusion models, the next two rows are video-only models, and finally the last ones are audio-only models. It is obvious that  the hybrid fusion strategy has superior performance compared to both the late and intermediate fusion strategies and also audio-only and video-only models. This is evident in its higher average validation accuracy. This can be attributed to its ability to leverage the strengths of both audiovisual fusion methods while minimizing their drawbacks. So, the multimodal model based on hybrid fusion strategy will be used in an interactive robot, whose name is Leaderbot, with security and supervisory functions to detect violence in public places such as airports. The relevant authorities will be informed whenever violence is detected in the area. 

\begin{table}[ht]
\centering
\caption{Comparison of the proposed method with other fusion , video-only, and audio-only models}
\label{tab:tab4}
\begin{tabular}{l|l|l|l|l}
\hline
\textbf{Model Name} & \textbf{ATA (\%)} & \textbf{ATL} & \textbf{AVA (\%)} & \textbf{AVL} \\ \hline
Proposed model & 97.63 & 0.08 & \textbf{96.67} & \textbf{0.12} \\
Intermediate Fusion & 97.33 & 0.09 & 93.78 & 0.14 \\ 
Late Fusion & 97.78 & 0.06 & 92.00 & 0.17 \\ 
\hline
I3D & 99.59 & 0.01 & 91.00 & 0.18 \\
Resnet(2+1)D \cite{Reinolds2022} & 90 & 0.08 & 89 & 0.13 \\
\hline
VGGish & 97.25 & 0.07 & 80.00 & 0.47 \\ 
Resnet 18 \cite{Reinolds2022} & 91 & 0.08 & 76 & 0.51\\ 
\end{tabular}
\end{table}

\subsection{Testing the best model}
\begin{figure}[ht]
\centering
\includegraphics[width=0.48\textwidth]{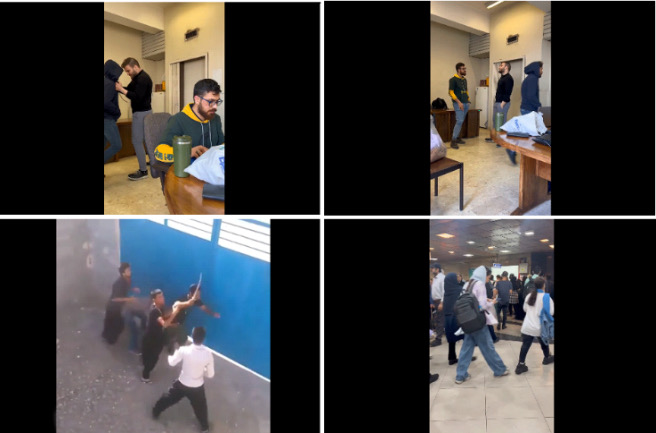}
\caption{. screenshots of the gathered dataset. (left column): Violent videos and (right column) non-violent videos.}
\label{fig:fig8}
\end{figure}

In order to test the performance of the proposed model on a real-world unseen data, a separate dataset is gathered by using a mobile phone camera, IPhone 13. This dataset consists of videos from various conditions and environments. The non-violent videos were recorded in crowded and quiet public places such as subways, streets, and universities campus. Meanwhile, the violent videos were captured from simulated and real conflicts. simulated conflicts are recorded with the help of students to simulate potential real-world scenarios. These simulations are designed to include both visual and auditory elements to create a more realistic portrayal of violent interactions. By applying the selected model to this diverse set of videos, the study aims to evaluate the effectiveness of the model in detecting real violent and non-violent scenes across different contexts. Fig. \ref{fig:fig8} displays frames from videos in the aforementioned dataset. It is necessary to mention that all the videos have the length of 5 seconds. A total of 54 videos were collected and applied to the model. Finally, the model succeeded in correctly recognizing 52 videos out of a total of 54 collected videos, which means an accuracy of about 96.29\%.

\section{CONCLUSION}
In this study, a novel hybrid fusion-based deep learning (HFBDL) approach is presented for human action recognition and violence detection. The proposed method surpasses audio-only, video-only, late fusion and intermediate fusion methods, attaining an impressive 96\% accuracy in violence detection. Its efficacy is demonstrated in real-world applications, including violence detection videos captured in public spaces. First of all, I3D and VGGish pretrained models were investigated and compared with other pretrained video and audio models to achieve best validation accuracy on the dataset. Then the selected pretrained video and audio models, I3D and VGGish, were used to extract high-level representation of features from the augmented dataset. The extracted features were fused using two main strategies of fusion, intermediate and late fusion, and also a combination of these two approaches, named hybrid fusion.

\begin{figure}[ht]
\centering
\includegraphics[width=0.30\textwidth]{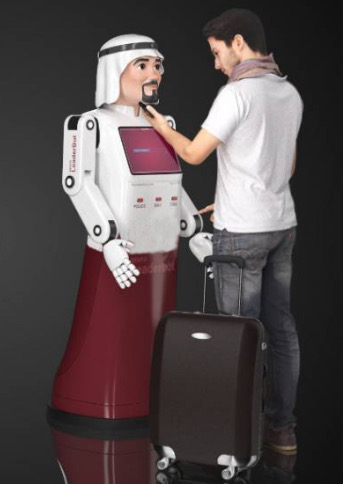}
\caption{Leaderbot - Designed in Distributed and Intelligent Optimization
Research lab}
\label{fig:fig9}
\end{figure}

The proposed HFBDL strategy demonstrated the best performance on both the augmented dataset and the gathered dataset. This model will be used in Leaderbot, which is shown in Fig. \ref{fig:fig9}, an interactive robot that is designed in Distributed and Intelligent Optimization Research lab.

Leaderbot has two main roles; environment monitoring (to detect violence) and guidance (to interact with people and answer their questions). Looking ahead, we intend to explore the implementation of a multimodal hybrid model with an attention mechanism in our future research. This approach should leverage both audio and visual modalities to potentially improve the classification of violent content. We hope that the attention mechanism will allow the model to selectively focus on the most relevant features of each modality, which could translate into better detection results across different contexts. Additionally, we plan to investigate whether the proposed multimodal model with attention mechanism can effectively utilize the entire RLVS dataset for training, despite the fact that some videos may lack sound or have irrelevant sound. By selectively focusing on the most relevant features of each  modality, we anticipate that the attention mechanism could help overcome this issue and potentially improve the overall robustness of the model.

\ifCLASSOPTIONcaptionsoff
  \newpage
\fi

\bibliographystyle{ieeetr}

\bibliography{library.bib}

\end{document}